\documentclass[english]{article}
\usepackage{eucal}
\usepackage[utf8]{inputenc}
\usepackage[T1]{fontenc}
\usepackage{babel}
\usepackage{amsmath}
\usepackage{amsfonts}
\usepackage{csquotes}
\usepackage{amstext}
\usepackage{amssymb}
\usepackage{graphicx}
\usepackage{comment}
\usepackage{subcaption}
\usepackage{fancyhdr}
\usepackage{siunitx}
\usepackage{hyperref}
\hypersetup{
    colorlinks=true,
    linkcolor=blue,
    filecolor=magenta,      
    urlcolor=cyan,
}
\usepackage{graphicx}

\usepackage{biblatex}
\begin{document}

\title{Unsupervised deep clustering and reinforcement learning can accurately segment MRI brain tumors with very small training sets}

\author{ \textbf{Joseph Stember}$^1$
\and 
\textbf{Hrithwik Shalu}$^2$}
\maketitle
\thispagestyle{fancy}
\noindent
\textsuperscript{1}Memorial Sloan Kettering Cancer Center, New York, NY, US, 10065 
\\
\textsuperscript{2}Indian Institute of Technology, Madras, Chennai, India, 600036
\\
\noindent
\textsuperscript{1}joestember@gmail.com
\\
\textsuperscript{2}lucasprimesaiyan@gmail.com 
\\

\begin{abstract}

\indent \textit{Purpose} Lesion segmentation in medical imaging is key to evaluating treatment response. We have recently shown that reinforcement learning can be applied to radiological images for lesion localization. Furthermore, we demonstrated that reinforcement learning addresses important limitations of supervised deep learning; namely, it can eliminate the requirement for large amounts of annotated training data and can provide valuable intuition  lacking in supervised approaches. However, we did not address the fundamental task of lesion/structure-of-interest segmentation. Here we introduce a method combining unsupervised deep learning clustering with reinforcement learning to segment brain lesions on MRI. 

\indent \textit{Materials and Methods}  We initially clustered images using unsupervised deep learning clustering to generate candidate lesion masks for each MRI image. The user then selected the best mask for each of $10$ training images. We then trained a reinforcement learning algorithm to select the masks. We tested the corresponding trained deep $Q$ network on a separate testing set of $10$ images. For comparison, we also trained and tested a U-net supervised deep learning network on the same set of training/testing images.

\indent \textit{Results}
Whereas the supervised approach quickly overfit the training data and predictably performed poorly on the testing set ($16 \%$ average Dice score), the unsupervised deep clustering and reinforcement learning achieved an average Dice score of $83 \%$.

\indent \textit{Conclusion}
We have demonstrated a proof-of-principle application of unsupervised deep clustering and reinforcement learning to segment brain tumors. The approach represents human-allied AI that requires minimal input from the radiologist \textit{without} the need for hand-traced annotation. 

\end{abstract}

\section*{Introduction}

Segmentation of lesions, organs, or other structures-of-interest is an integral component of artificial intelligence (AI) in radiology \cite{mcbee2018deep,saba2019present,mazurowski2019deep}. Essentially all AI segmentation, like the tasks of localization and classification, falls within the category of supervised deep learning. 

A supervised deep learning research project focusing on segmentation typically begins by accruing and pre-processing a large number of appropriate images. As a general rule, hundreds of images are required for successful training. Next, the process of annotation begins, which entails radiologists or other imaging researchers tracing outer contours around each structure-of-interest, thereby creating a mask. Once enough masks have been acquired, the data is typically "augmented" via a series of rotations, scaling, translations, and/or the addition of random pixel noise to artificially produce a larger training set. At this point, the data set is divided between a training/validation set, which comprises the vast majority of the image data set, and a smaller separate testing set. A convolutional neural network (CNN) is trained by repeated forward and backpropagation through the CNN with the training set images as input and output compared to masks for the loss. The two most common CNN architectures for segmentation are mask-regional-CNN (mask-RCNN) \cite{he2017mask} and U-net \cite{ronneberger2015u}.

We sought to address three key shortcomings in current supervised deep learning approaches:

\begin{enumerate}
  \item Requirement of large amounts of expert-annotated data. This can be expensive, tedious, and time-consuming. For example, it has been estimated that segmenting $1,000$ images may require a month of full-time work by two experts \cite{chartrand2017deep}. 
  \item Lack of generalizability, making the algorithm “brittle” and subject to grossly incorrect predictions when even a small amount of variation is introduced. This can occur when applying a trained network to images from a new scanner, institution, and/or patient population \cite{wang2020inconsistent,goodfellow2014explaining}. 
  \item Lack of insight or intuition into the algorithm, which limits the confidence needed for clinical implementation and restricts potential contributions from non-AI experts with advanced domain knowledge (e.g., radiologists or pathologists) \cite{buhrmester2019analysis,liu2019comparison}.
\end{enumerate}

In recent work \cite{stember2020deep,stember2020reinforcement}, we introduced the concept of radiological reinforcement learning (RL): the application of reinforcement learning to analysis of medical images. We showed that RL can address these challenges by applying RL to lesion localization. 

However, segmentation can provide additional information such as lesion volume, which is particularly useful for follow-up determining radiological response to treatment. Here we apply RL to the task of segmentation. In order to do so, we leverage another important, though less prevalent, branch of AI in radiology: unsupervised deep learning. Unsupervised deep learning performs data clustering, grouping the data sets into different classes. The identity of these classes is left to human users to determine, thereby providing post-training "supervision." 

Hence we proceed in two major phases: 
\begin{itemize}
    \item Use unsupervised clustering CNN or unsupervised deep clustering (UDC) to generate clusters that are candidate lesion masks. The user then selects the cluster that serves as the best mask. Then for this proof-of-principle application, we simplify by reducing the tiling of each image to two clusters, the user-selected mask and its complement, for the next step in the training set.
    \item Using a training set comprised of lesions and masks obtained from UDC, train a deep reinforcement learning (RL) deep Q network (DQN) in tandem with Q-learning to predict the best cluster to serve as lesion mask.
\end{itemize}

Using this method, we produce lesion segmentations with minimal input from the user. The user is spared from manually tracing out a single lesion border. Additionally, given the data efficiency of RL, even the user effort to select masks from the UDC step can be minimized. This is due to the ability of RL to train effectively on very small data sets, as we have seen in our recent work \cite{stember2020deep,stember2020reinforcement}. We will show in this paper that we were able to produce accurate lesion segmentations with only $10$ user-directed mask selections.

\section*{Methods}

\subsection{Overview}

We can divide our approach into five key steps:
\begin{itemize}
\item Collect $20$ glioma MRI 2D images. The first $10$ will be used for the training set, and the remaining $10$ will comprise the testing set.
\item Use an unsupervised deep clustering (UDC) network to generate a set of possible lesion masks.
\item User selects appropriate mask.
\item As above, obtain $10$ training set images $+$ masks.
\item Train a reinforcement learning (RL) algorithm using $Q$ learning and a deep $Q$ network to select the correct lesion masks automatically. 
\end{itemize}

\subsection{Data collection}

We collected $20$ publicly available T2 FLAIR images of gliomas as screen captures using web sites accessed via Google Images \cite{web_1,web_2,web_3,web_4,web_5} as well as published articles \cite{smits2015neurological,radbruch2011relevance,da2011pseudoprogression,zhang2017clinical,wen2017response}. As we used only publicly available images, IRB approval was deemed unnecessary for this study.

\subsection{Clustering}

\subsubsection{Superpixel generation}

Consider for example one of the training images. We start by generating a tiling of our image using the SLIC method \cite{yang2016joint,achanta2012slic}. This approach separates the image into $N_{sp}$ superpixels using nearest-neighbors with the features of red-green-blue color channel pixel intensity, each in the range of $0$ to $255$ and $x,y$ position. This tends to cluster regions of the image that are of similar color and are spatially close. An example superpixel tiling is displayed on the right image in Figure \ref{fig:target_vs_prediction}. In order to ensure small enough superpixels so as not to artifactually group together disparate regions, we used a high number of superpixels $N_{sp} = 1 \times 10^4$. The python function $skimage.segmentation.slic$ produces the tiling ultimately with a number of superpixels close to $N_{sp}$. We set a parameter called compactness equal to $100$ to balance the influence of color proximity and space proximity. The SLIC tiling forms a scaffolding for subsequent steps. 

\subsubsection{Clustering CNN}

\begin{figure}
\centering
\includegraphics[width=11.5cm,height=8.5cm]{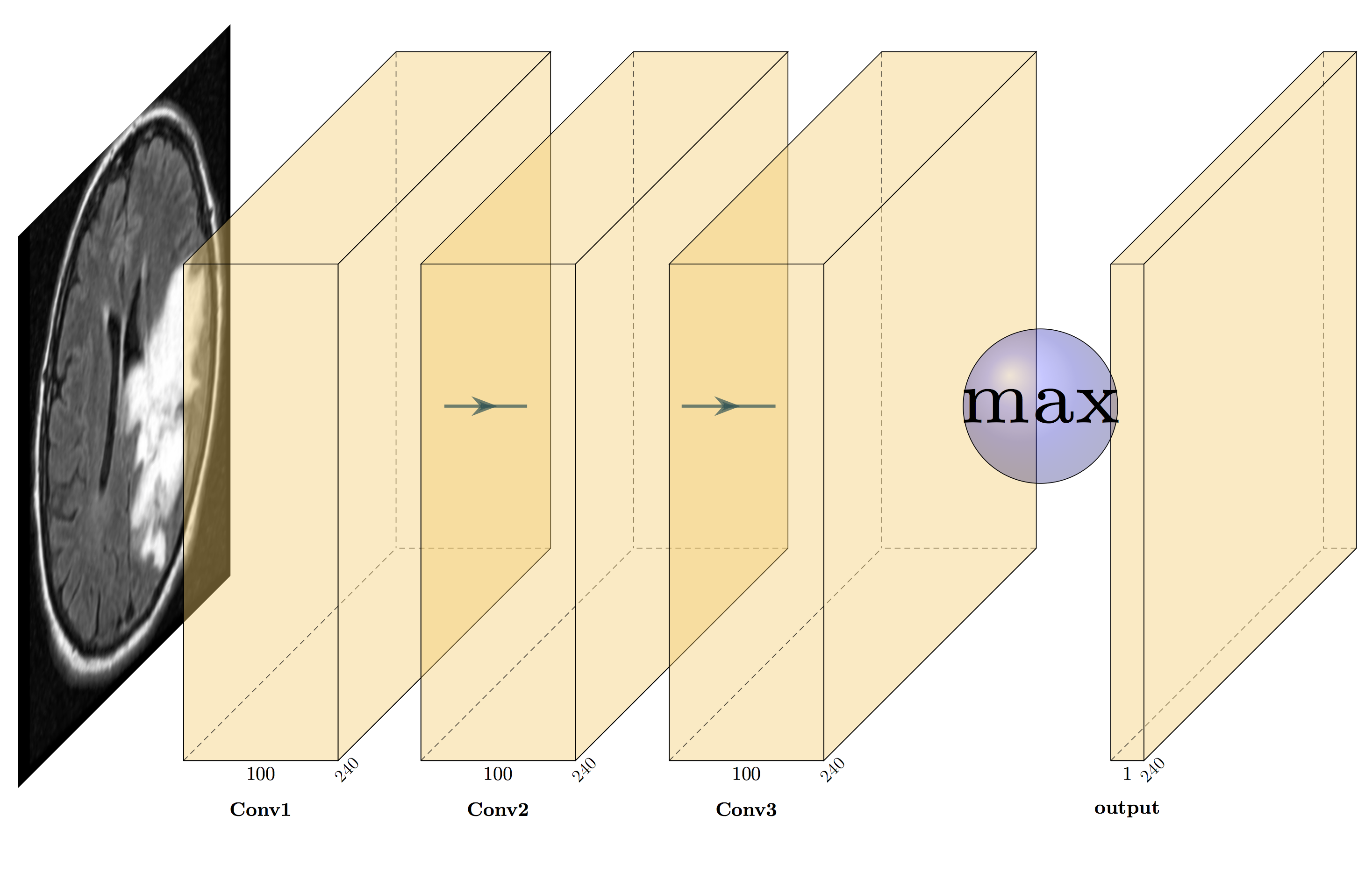}
\caption{Schematic of the unsupervised clustering convolutional neural network. The raw image is fed as input into the network with convolutional layers via $3 \times 3$ kernals and zero padding to keep subsequent layers the same $H \times W$ size. The three convolutional layers have $100$ channels and each pixel has $100$ features. Then $argmax$ is obtained over the $100$ features for each pixel, selecting the most important feature over the image.  }
\label{fig:clustering_network_architecture}
\end{figure}

\begin{figure}
\centering
\includegraphics[width=12cm,height=10cm]{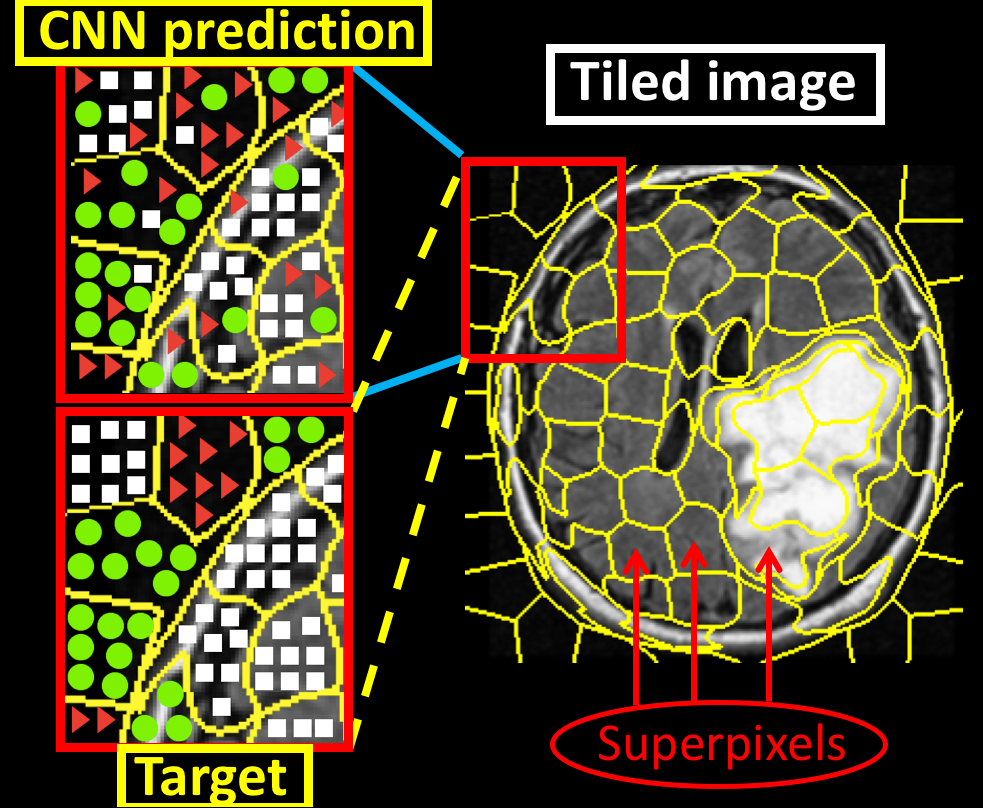}
\caption{Illustration of target vs. output prediction from the network illustrated in Figure \ref{fig:clustering_network_architecture}. On the right is a tiling of the input image. The individual cells are the so-called superpixels, generated by the SLIC method. For the purposes of illustration, a small subset of the image is analyzed on the left panel. The CNN from Figure \ref{fig:clustering_network_architecture} predicts most important features of the $100$-feature embeddings in the penultimate layer of said figure. The goal is for each pixel in a given superpixel to be the one type of feature. Here we illustrate three features as three shapes. The algorithm to generate the target is to make all pixels in a given superpixel equal to the most numerous or common feature type predicted in the CNN output.  }
\label{fig:target_vs_prediction}
\end{figure}

We run a forward pass of the convolutional neural network (CNN) shown in Figure \ref{fig:clustering_network_architecture}. Part of a sample output of the forward pass is displayed as the upper left panel in Figure \ref{fig:target_vs_prediction}. The target is produced from this output by making the most numerous features selected for each superpixel the only feature present for each pixel within that superpixel. The target obtained from the sample output of Figure \ref{fig:target_vs_prediction} is shown as the lower left image of Figure \ref{fig:target_vs_prediction}. 

Then cross entropy between target and output provides pixel-wise classification CNN loss, which we minimize via backpropagation. The CNN thus learns during training to predict features for which all or as many as possible pixels in each superpixel have the same value. In other words, as the network trains, it become more confident about the categories to which small regions (superpixels) in the image belong. Then, grouping the superpixels together provides us with larger image clusters. Ours is thus an agglomerative ("bottom up") hierarchical image clustering. 

Because the network in Figure \ref{fig:clustering_network_architecture} is unsupervised, it does not know the category or label of the clusters. However, it does learn that certain tiny regions belong together in the same cluster and class. Then it is up to the user to assign actual identities to these regions. A key step in our approach is to minimize the burden of user or expert oversight needed to inform the unsupervised network, thereby guiding it toward more meaningful predictions.

Our network, as shown in Figure \ref{fig:clustering_network_architecture}, consists of three convolutional layers. We use $3 \times 3$ kernals with $NC = 100$ channels and zero padding to maintain the $H \times W$ dimensions of the convolutional layers. Hence these layers are of size $H \times W \times NC = 240 \times 240 \times 100$. Because the third layer also has these dimensions, the $NC$ channel represents a $100$-CNN-feature embedding. This provides higher order features than the simpler red-green-blue and $x,y$ coordinates, and the most important features for clustering are determined by backpropagation. As stated above, loss is calculated as the cross entropy between CNN output and target. Since the target for each superpixel can be specified via one hot encoding vector, the closer all the CNN output pixels in the superpixel are to being of the same class and the class matching the one hot encoding vector, the higher the $\log$ and the lower the negative sum of $\log$ values, thus the lower the loss.

Training the clustering CNN is somewhat unique in that it ends when the number of clusters/classes, $N_{cl}$, reaches  25. The clustering CNN trains to make output pixel features more homogeneous within superpixels, but also across superpixels. This represents the agglomerative building up of increasingly larger and "higher level" clusters, starting from the tiny superpixels (roughly $240 \times 240 \text{ pixels} \div 10,000 \text{ superpixels} \approx 6 \text{ pixels per superpixel} $) and building up to what would ultimately be $N_{cl} = 1$ (all of the image being in one class). Then it is up to the user to determine the optimal number of classes at which to stop the training. Based on trial and error, we found that $N_{cl}=25$ is a good number of classes that produces masks that encompass important structures, including the lesions. Hence, we set training to conclude as soon as $N_{cl}$ reaches $25$. Training to the goal of $25$ classes typically takes on the order of $20$ epochs, comprising a few seconds of training time.

Since we want the user to interactively oversee the selection of lesion masks and to reduce the need to wait around for very long, we want the network to train almost instantaneously. As such, candidate lesion masks can be provided by the CNN from Figure \ref{fig:clustering_network_architecture} and the user can select the correct masks in real time. In order to achieve fast training, we use the relatively large learning rate of $0.1$. However, with this higher learning rate, we must mitigate the risk that the tapering off in $N_{cl}$ during training could occur too quickly, or "skip" over intermediate values. Were this to occur, intermediate-sized features in the images might not be adequately factored into the ultimate clustering, comprising a loss of valuable information. In order to counteract the possibility of skipping over intermediate-sized clusters, we apply batch normalization after each convolutional layer. This keeps the network weights from growing too large. We also use stochastic gradient descent as the optimization iterative method with the standard momentum value of $0.9$. Unlike other optimization algorithms, stochastic gradient descent has no acceleration, which could combine with the large learning rate to make training unstable.  

\subsubsection{Evaluating clusters and selecting best masks}

For each image, having generated $N_{clusters}$, we remove small candidate masks that are likely artifactual. We do so by requiring that all candidate masks be over $1 \times 10^3$ pixels large. Furthermore, to avoid candidates that are overly large, we include an upper size boundary, which is the total number of pixels in the image minus $1 \times 10^3$ pixels. We then display all candidate masks for the viewer to select the best. This is displayed in Figure \ref{fig:mask_selection}. In that case, for example, the user would instantly recognize that the first image contains the best mask. We imagine that in future implementations users will click on or dictate the best image in a fluid and efficient process. However, for this proof of principle, the user recorded the best image number by entering it into a python array. In our proof of principle, we use the center of the mass point for each mask as this fiduciary point. Any point, even one selected at random from the image, could serve as this fiduciary point. 

We can imagine the user clicking on the best cluster for lesion mask selection. This would provide for the recording of a point within the lesion. We use this as a fiducial point $p_{\text{fid}}$.  

Proceeding as above for each of the $10$ training set images, we obtain an array of length $10$ containing indices of the best mask selections for each image. Then we are able to re-produce the $10$ best masks to accompany the raw images by calling the array. Having obtained the $10$ training masks, we now turn to the task of training a separate network to detect and segment lesion masks without any user interaction.

\subsection{Reinforcement learning for lesion segmentation}

\subsubsection{Reinforcement learning environment and definition of states, actions, and rewards}

\paragraph{State space}

\begin{figure}
\centering
\includegraphics[width=12cm,height=12cm]{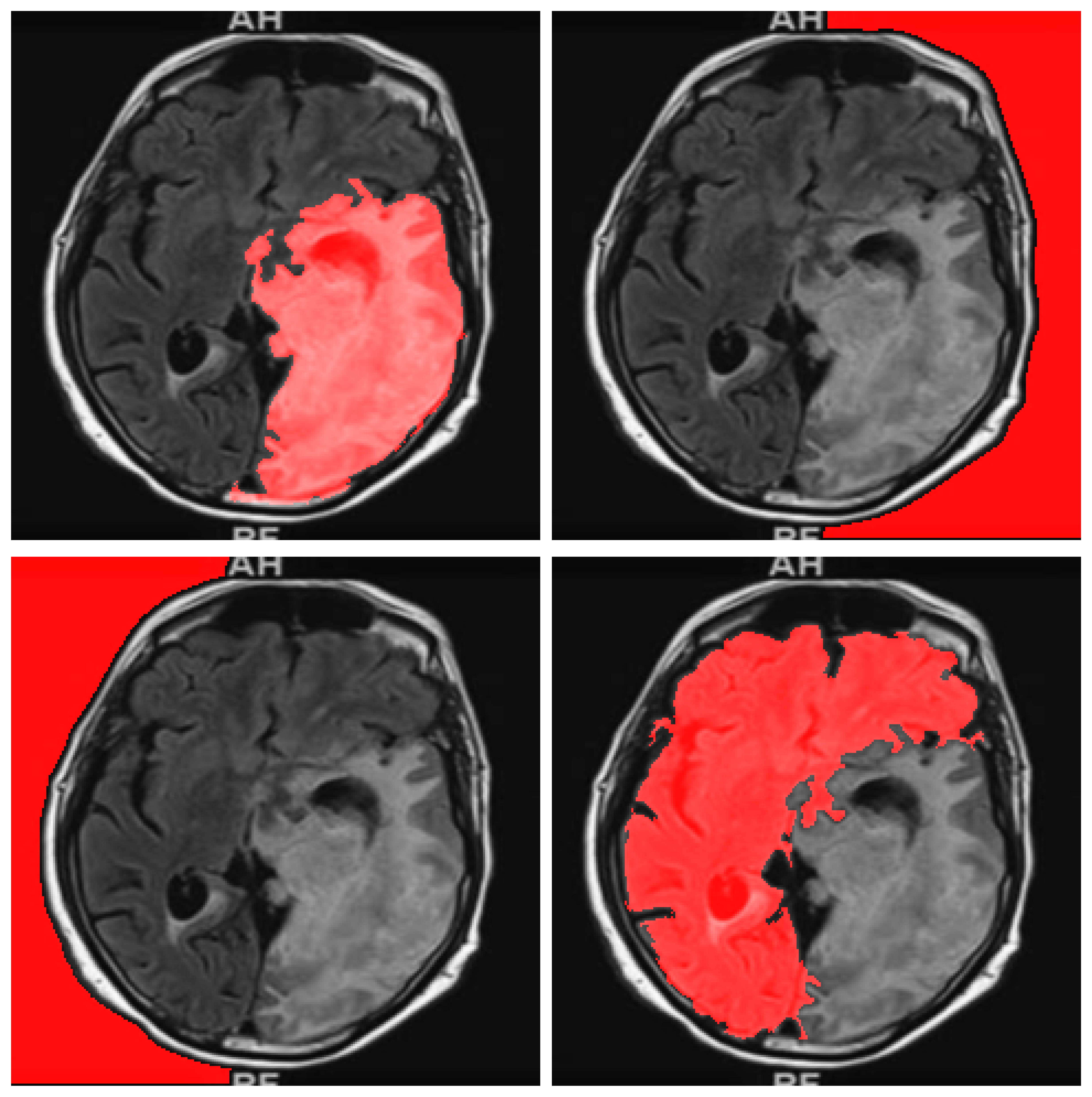}
\caption{Set of clusters produced by the unsupervised deep clustering CNN. These serve as lesion mask candidates. The user can select the best image or mask number; here the best mask is clearly the first one, in the top left image. That cluster $\mathcal{M}$ and its complement $\mathcal{M}^C$ are then used to train the reinforcement learning network. The reinforcement learning network learns to select which is the best cluster (mask or its complement) to serve as lesion mask. }
\label{fig:mask_selection}
\end{figure}

Here the various states are the different clusters as produced by the unsupervised CNN, as in Figure \ref{fig:clustering_network_architecture}. For example in Figure \ref{fig:mask_selection}, the four possible states are the four sub-figures. The state that corresponds to the correct lesion mask is the first (upper left) sub-figure. In general, for an image tiling consisting of $N_{cl}$ clusters, there are $N_{cl}$ different states. For the sake of simplicity in this proof-of-principle, we restrict the state space to two possible clusters with a cluster overlaying the mask $\mathcal{M}$ and its complement $\mathcal{M}^C$. 

As constructed thus far, our scenario fits the framework of a standard multi (two)-armed bandit problem \cite{sutton2018reinforcement}. This being the case, we would not have the multi-step sequence of states ${ s_i }$, which would be required to employ the Bellman Equation for updating $Q$. We thus allow for five subsequent states to be explored per episode of training. As such, five transitions can be stored in the replay memory buffer of each episode. 

\begin{figure}
\centering
\includegraphics[width=12cm,height=12cm]{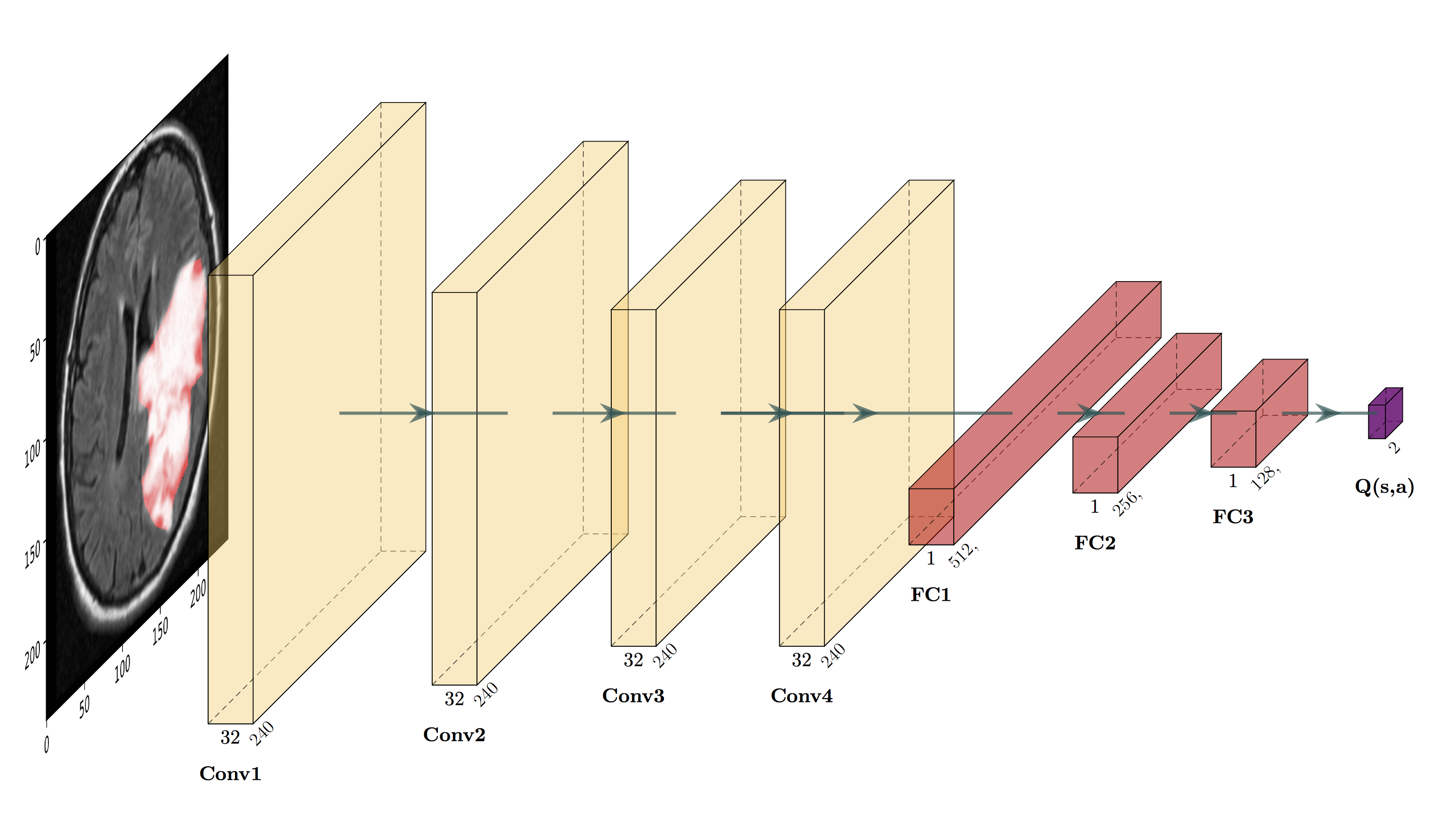}
\caption{Architecture of deep $Q$ learning CNN to select which region should be the lesion mask. }
\label{fig:DQN_architecture}
\end{figure}

As in Figure \ref{fig:mask_selection}, we overlay color-coded clusters onto our grayscale images to represent states $s_t$. The initial input image $s_1$ shows the lesion mask colored in red, as shown in Figure \ref{fig:DQN_architecture}. For subsequent steps of environmental sampling for $Q$ learning, we can in general at time $t$ define the state $s_t$ by:

\begin{equation} 
s_t =  \begin{cases}
        \text{red mask} & \text{if } a_{t-1} = 1 \\
        \text{green mask} & \text{if } a_{t-1} = 2 \text{,}
    \end{cases}
\label{state_eqn}
\end{equation}
where $a_{t-1}$ is the action taken in the previous step.

\paragraph{Action space}

The actions our agent can take consist of selecting the cluster in the image tiling that represents the lesion mask. In general, there are $N_{cl}$ possible actions. In the simplified environment we evaluate in the present work, $N_{cl}=2$ so that the action represents selection of either $\mathcal{M}$ or $\mathcal{M}^C$. More specifically, the action selects a cluster as lesion mask by selecting the region to which $p_f$ belongs. For the training set images, since we selected the center of mass as the points of interest, these are all inside the lesion masks.

In other words, the action space is generally $\mathcal{A} \in \mathbb{N}_{0}^{N_{cl}}$, but in our simplified case $\mathcal{A} \in \mathbb{N}_{0}^{2}$ is defined by:

\begin{equation} 
\mathcal{A} = \begin{pmatrix} 1 \\ 2 \end{pmatrix} = \begin{pmatrix} \text{predict } p_f \in \mathcal{M} \\ \text{predict } p_f \in \mathcal{M}^C \end{pmatrix} \text{.} \label{action_eqn}
\end{equation}

\paragraph{Reward structure}

We seek to reward and incentivize choosing the correct cluster while penalizing incorrect cluster selection. The reward scheme we use is thus given by:
\begin{equation} 
r_t =  \begin{cases}
        +1 & \text{if } a_{t-1} = 1 \text{ and point is within the lesion} 
        \\ +1 & \text{if } a_{t-1} = 2 \text{ and point is outside the lesion}
        \\ -1 & \text{if } a_{t-1} = 1 \text{ and point is outside the lesion}
        \\ -1 & \text{if } a_{t-1} = 2 \text{ and point is within the lesion}
        \text{.}
    \end{cases}
\label{reward_eqn}
\end{equation}

\subsubsection{Training: Deep $Q$ network}

As in our recent work \cite{stember2020deep,stember2020reinforcement}, we use a convolutional neural network (CNN) that we term a Deep $Q$ network (DQN) to approximate the function for $Q_t(a)$. The architecture of our DQN is displayed in Figure \ref{fig:DQN_architecture} and is very similar to DQNs used in our recent work. As before, it employs $3 \times 3$ kerns with stride of $2$ and padding such that resulting filter sizes are unchanged. There are four convolutional layers, following up exponential linear unit (elu) activation functions. The last convolutional layer is followed by fully connected layers, ultimately producing a two-node output. The two output nodes correspond to the set of two $Q(s,a)$ values, which depend on the state $s$ and the two possible actions $a_t={1,2}$. $Q(s,a)$ is well-known from reinforcement learning as the action value function. We use mean squared error loss with batch size of $n_{\text{batch}}=16$ and learning rate of $1 \times {10}^ {-4}$. 

Our DQN loss is the difference between output $Q$ values, $Q_{\text{DQN}}$, and the “target” $Q$ value, $Q_{\text{target}}$. The former is computed by a forward pass, $F_{\text{DQN}}(s_t)$ of the DQN, $Q_{\text{DQN}}^{(t)}=F_{\text{DQN}}(s_t)$. The latter is computed by sampling the environment via the Bellman equation / temporal difference learning, as below. 

\subsubsection{Q learning via $TD(0)$ temporal difference learning}

As training learns the functional approximation of $Q$, bringing $Q_{\text{DQN}}$ closer to $Q_{\text{target}}$, we wish to optimize the latter toward the best possible $Q$ value, $Q^{\star}$, by sampling from the environment. As in our recent work, we achieve this via temporal difference $Q$ learning in its simplest form: $TD(0)$. With $TD(0)$, we can update $Q_{\text{target}}^{(t)}$ by way of the Bellman Equation:
\begin{equation}
    Q_{\text{target}}^{(t)} = r_t + \gamma max_a Q(s_{t+1},a) \text{,}
\label{bellman_eqn}
\end{equation}
where $\gamma$ is the discount factor and $max_a Q(s_{t+1},a)$ is equivalent to the state value function $V(s_{t+1})$. The most important part of the environment sampled is the reward value $r_t$. 

Over time, with this sampling, $Q_{\text{target}}^{(t)}$ converges toward the optimal $Q$ function, $Q^{\star}$. In our implementation, for each episode, the agent was allowed to sample the image for $20$ steps. We set $\gamma = 0.99$, a frequently used value that, being close to $1$, emphasizes current and next states but includes those further in the future. 

$Q$ learning is off-policy in the sense that the policy for sampling state-action space is not the same as the policy followed to select new actions. As in our recent work \cite{stember2020deep,stember2020reinforcement}, we select each action $a_t$ at time step $t$ according to the off-policy epsilon-greedy algorithm, which seeks to balance exploration of various states by exploiting known best policy, according to:
\begin{equation}
    a_t =
    \begin{cases}
        max_{a \in A} \{ Q_t ( a ) \} & \text{with probability $\epsilon$} \\
        \text{random action in $A$} & \text{with probability $1-\epsilon$} \text{.}
    \end{cases}
\end{equation}
for the parameter $\epsilon < 1$. We used an initial $\epsilon$ or $0.7$ to allow for adequate exploration. As $Q$ learning proceeds and we wish to increasingly favor exploitation of a better known and more optimal policy, we set $\epsilon$ to decrease by a rate of $1 \times 10^{-4}$ per episode. The decrease continued down to a minimum value $\epsilon_{\text{min}}=1 \times 10^{-4}$, so that some amount of exploring would always take place. 

\subsubsection{Replay memory buffer}

As in our previous work \cite{stember2020deep,stember2020reinforcement}, and as per standard $Q$ learning \cite{sutton2018reinforcement}, we store states, actions taken, next states, and rewards from the actions as transition values. More formally, for time $t$, we store the given state $s_t$, action $a_t$, resulting in reward $r_t$ and bringing the agent to new state $s_{t+1}$. We collect these values in a tuple, called a transition, $\mathcal{T}_t=(s_t,a_t,r_t,s_{t+1})$. For each successive time step, we stack successive transitions as rows in a transition matrix $\mathbb{T}$ up to a maximum size of $N_{\text{memory}}=1,800$ rows. This is the replay memory buffer, which allows the DQN to learn from past experience sampling from the environments of the training images and states. After reaching $N_{\text{memory}}=1,800$ rows, $\forall t > N_{\text{memory}}$, new rows $\mathcal{T}_t$ are added while earlier rows are removed from $\mathbb{T}$. 

During DQN training, batches of batch size $n_{\text{batch}}$ transitions are randomly sampled from $\mathbb{T}$. This allows a thorough and well distributed sampling of environments and states so that $Q_{DQN}$ is as generalized as possible. 

In all, we sample for $N_{\text{episodes}}=300$ episodes, each consisting of five successive states, each of which begins with $s_1$, which is randomly selected for each episode among the $10$ training sets.

\section{Results}

\subsection{Application of trained UDC and RL to testing set}

For each of the $10$ testing set images, we predict to which cluster ($\mathcal{M}$ or $\mathcal{M}^C$) our fidelity point $p_f$ belongs. In this case, for simplicity, we used the lesion mask center of mass points. For each image, we generate a state as we did for the training images. Then we run each state through the trained DQN and extract the best action as the index of the larger of the two predicted $Q$ values, i.e., $argmax(Q_{\text{DQN}})$. This selects the predicted lesion mask.

\subsection{Training a U-net for comparison}

As in our previous work, we seek to compare the performance of RL to that of supervised deep learning \cite{stember2020deep,stember2020reinforcement}. As the task here is lesion mask generation/segmentation, the supervised CNN we use is the U-net architecture that we previously applied successfully to segment brain aneurysms \cite{stember2019convolutional} and meningiomas \cite{stember2019eye}. 

We trained the $16$-layer U-net on our training set of $10$ images and corresponding masks for $50$ epochs. We use epoch in analogy to the episodes from RL. We used a batch size of four, Adam optimizer with learning rate of $1 \times 10^{-5}$, and loss function given by the negative Dice similarity score between network output and hand annotated mask. 

These parameters and network architecture provided accurate segmentation in our prior work when provided with training set sizes in the hundreds, further increasing by data augmentation \cite{stember2019convolutional,stember2019eye}. However, in this case, given the extremely small training set size, the network began overfitting the training data early in the training process and badly overfitted the training set, an unsurprising result. The result was not generalizable to the separate testing set, for which the Dice score was $16 \%$.  

\subsection{Comparison between unsupervised deep clustering and reinforcement learning segmentation versus supervised deep learning/U-net}

\begin{figure}
\centering
\includegraphics[width=12cm,height=9cm]{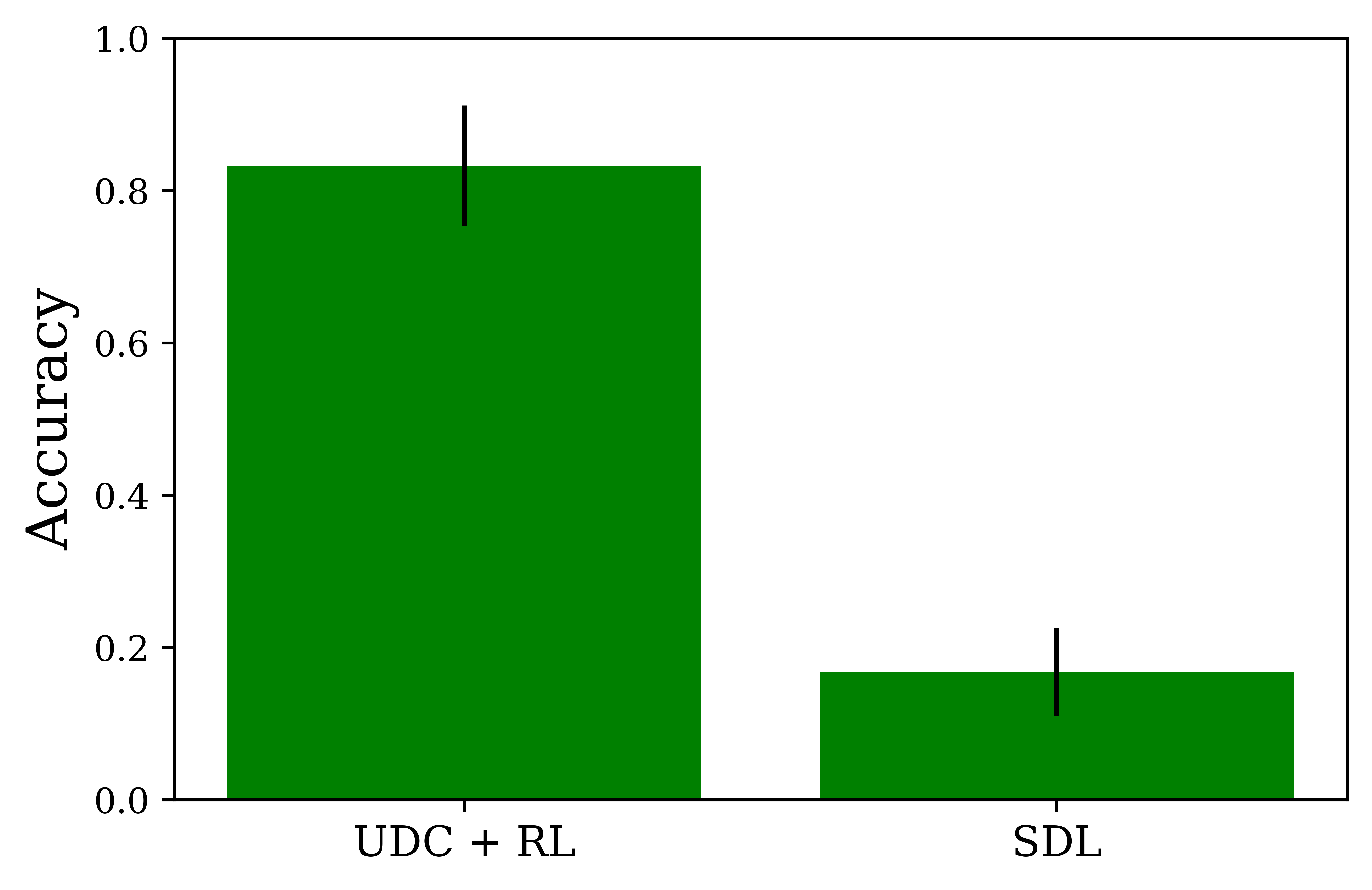}
\caption{Bar plot comparison between testing set accuracy (measured as Dice score) for unsupervised deep clustering and reinforcement learning map selection ($UDC + RL$) and supervised deep learning/U-net ($SDL$). The average value is given by the height of the bars, and the standard deviation is represented by the error bars. }
\label{fig:acc_comp}
\end{figure}

The average Dice similarity score of UDC and RL segmentation was $83\%$, while that of the trained deep supervised network with U-net architecture was $16\%$. The difference was statistically significant with a $p$-value of $5.3 \times 10^{-14}$. A visual comparison of the performance of the two trained networks is shown as a box plot in Figure \ref{fig:acc_comp}. 

\section{Discussion}

We have shown that a combination of unsupervised deep clustering and deep reinforcement learning can produce accurate lesion segmentations. Furthermore, it is able to do so with a very small training set, similar to results we saw in recent lesion localization work with RL \cite{stember2020reinforcement,sutton2018reinforcement}. Even with a small training set, the user was only required to select the best mask from the candidates produced by the unsupervised clustering CNN. The user did perform a single contour tracing annotation.

This was an initial proof-of-principle work with some important limitations and goals for future work. For simplicity, we restricted the number of clusters produced by the clustering CNN to $N_{cl}$ 2 clusters/classes, the lesion mask, and its complement. In general, the clustering CNN would tile the image into $N_{cl}$ clusters. Through trial and error, we found that for these images, around $25$ clusters gave the best segmentation of important structures. Our general approach would be a slight generalization of that presented here, using $N_{cl}$ possible actions, which would select the cluster of interest. The cluster ultimately selected would be the predicted mask. The initial tiling into superpixels by SLIC was just one of many initial schemes that can be incorporated into a CNN-based clustering algorithm. Others may prove more effective for the types of images we are analyzing, and a comparison between different approaches is a topic for future work. 

Finally, our ultimate goal is extension of this approach to fully three-dimensional image stacks. This would provide lesion volumes, as opposed to the two-dimensional areas. 

\section*{Conflicts of interest}

The authors have pursued a provisional patent based on the work described here.

\begin{figure}
\centering
\includegraphics[width=11.5cm,height=8.5cm]{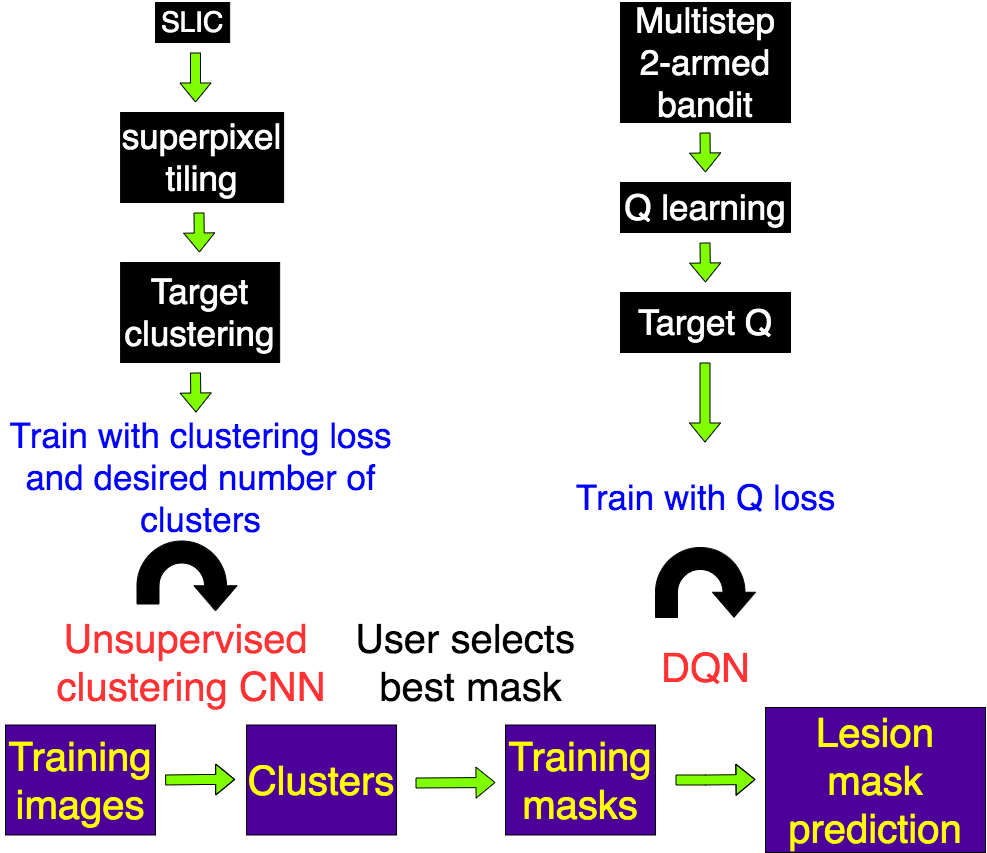}
\caption{Overview of the training scheme.  }
\label{fig:overview_flow_chart}
\end{figure}

\begin{figure}
\centering
\includegraphics[width=12cm,height=12cm]{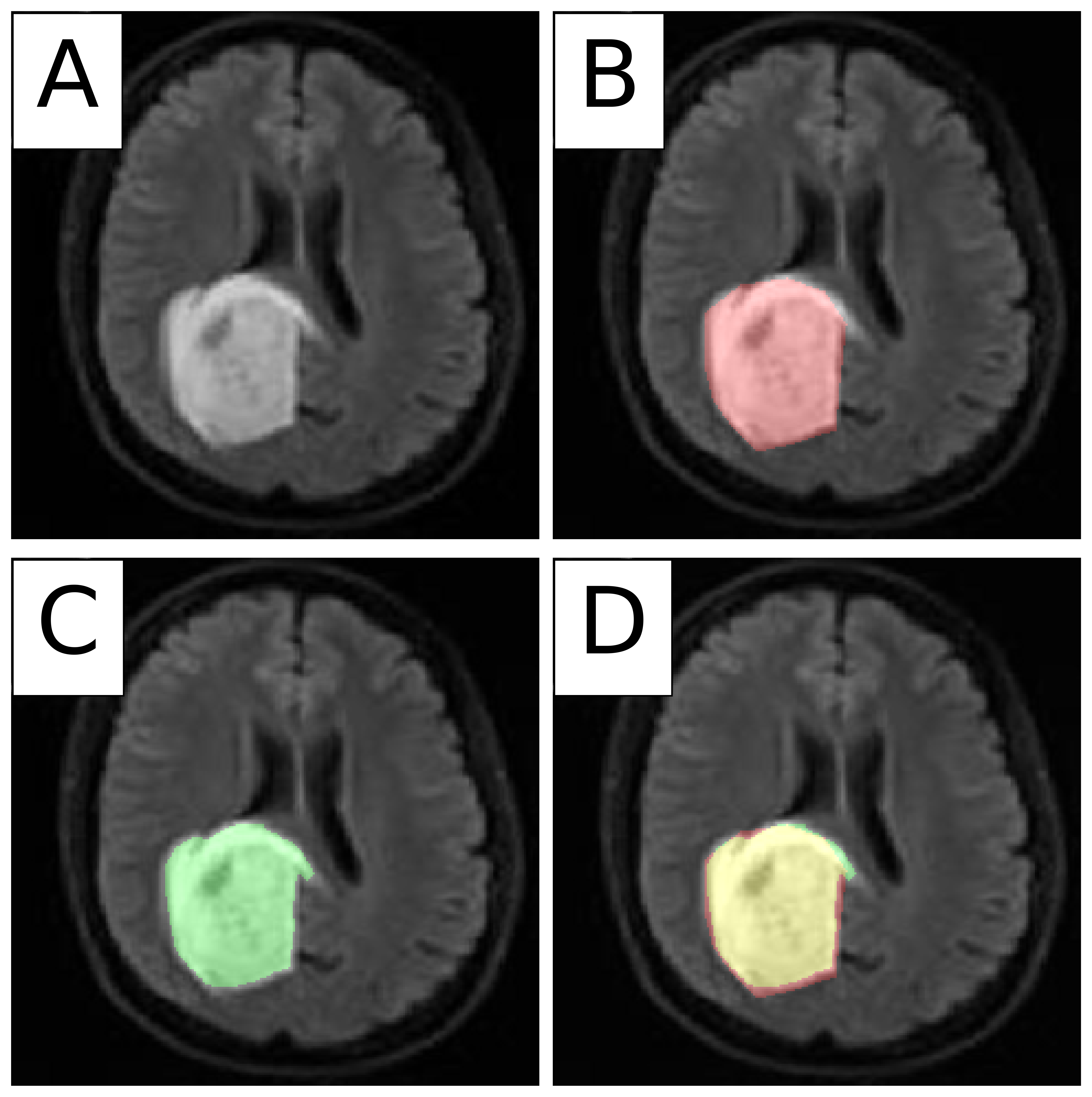}
\caption{Sample mask predictions. (A) shows one of the testing set images, (B) displays the hand annotated mask overlaid in red, (C) shows the predicted mask from the unsupervised deep clustering reinforcement learning approach overlaid in green, and (D) shows both masks overlaid, noting that their region of overlap becomes yellow.}
\label{fig:mask_example}
\end{figure}

\begin{figure}
\centering
\includegraphics[width=12cm,height=9cm]{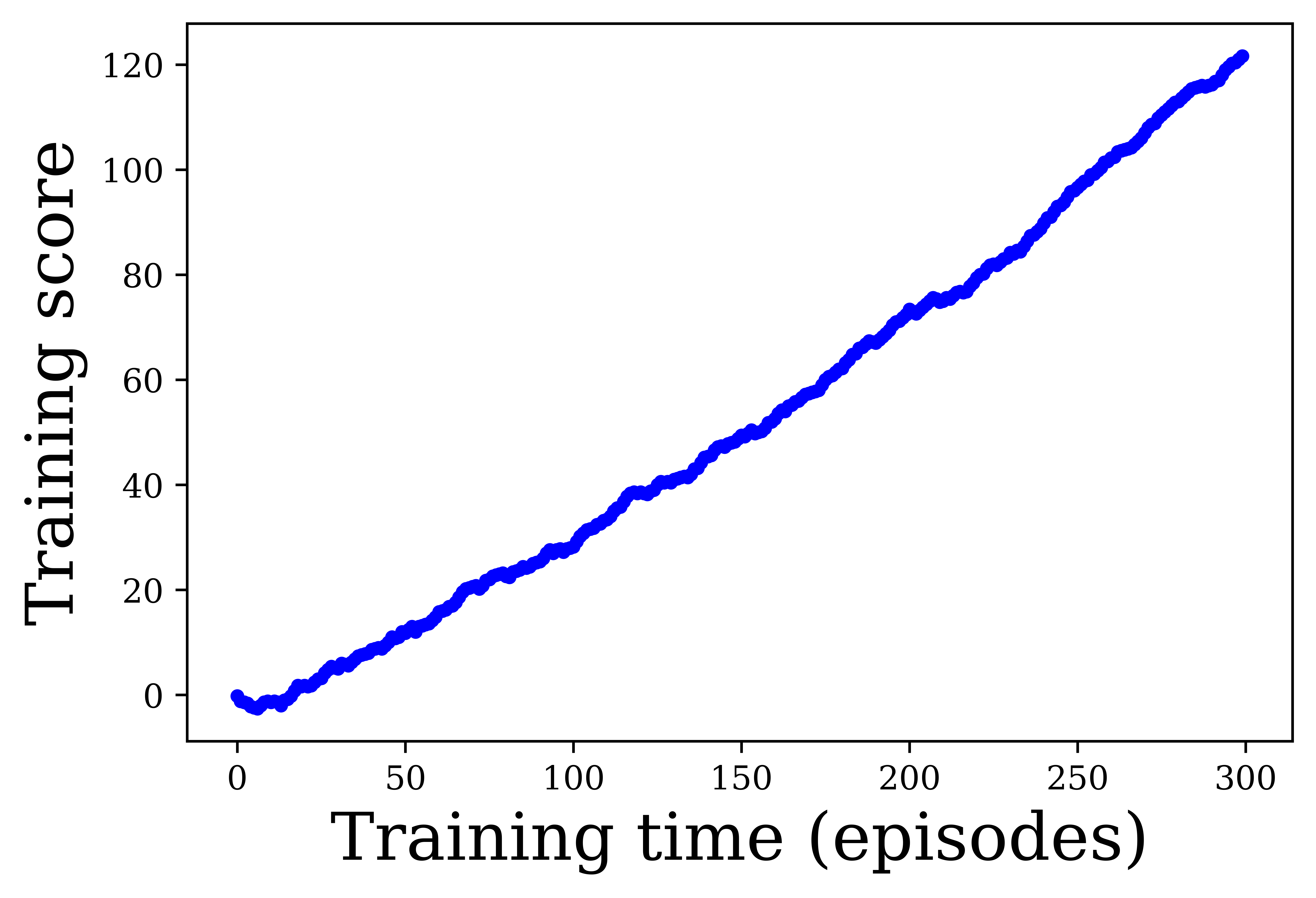}
\caption{Training set accuracy as a function of training time. A steady and monotonic increase is manifest.}
\label{fig:training_score}
\end{figure}

\begin{figure}
\centering
\includegraphics[width=12cm,height=9cm]{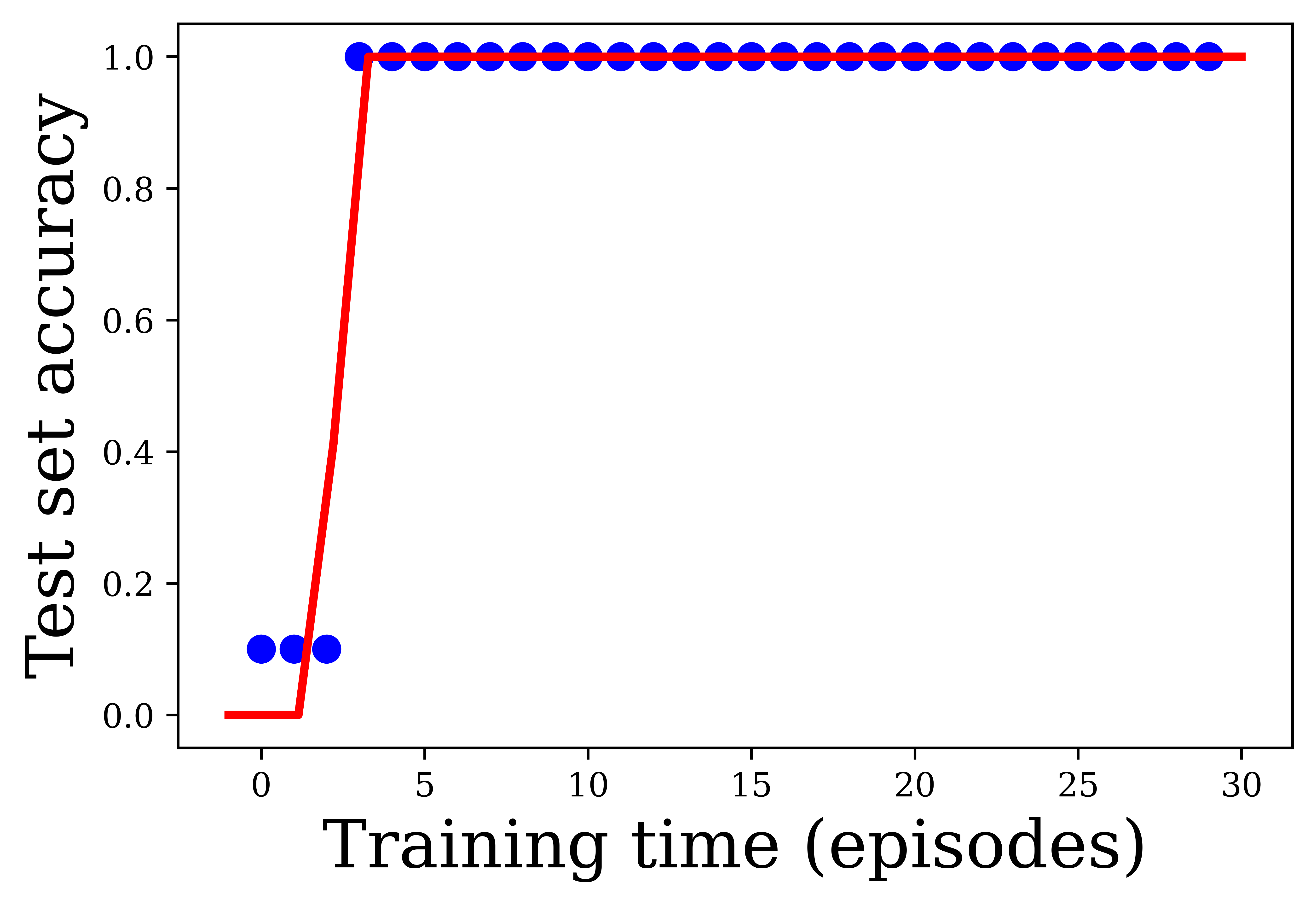}
\caption{Test set accuracy of reinforcement learning map selection as a function of training time. All $10$ test set images have the correct mask selected by the fourth episode and there is no subsequent dip in accuracy to suggest over-fitting. Best fit sigmoid function curve is shown in red.}
\label{fig:testing_score}
\end{figure}

\printbibliography

\end{document}